\RequirePackage{fix-cm}
\documentclass[smallextended]{svjour3}       
\smartqed  
\usepackage{color}
\usepackage{graphicx}
\usepackage{amsmath}
\usepackage{cite}
\usepackage[colorlinks=true,linkcolor=blue,citecolor=blue]{hyperref}
\begin{document}

\title{A statistical reduced-reference method for color image quality assessment}


\author{Mounir Omari \and  Mohammed El Hassouni \and Abdelkaher Ait Abdelouahad  \and Hocine Cherifi
}


\institute{M. Omari and M. El Hassouni\at
              LRIT URAC 29, University Mohammed V-Agdal\\
              \email{mouniro870@gmail.com, mohamed.elhassouni@gmail.com}           
           \and
           A. Ait Abdelouahad \at
              University Ibn Zohr\\
              Agadir, Morocco.\\
              \email{a.abdelkaher@gmail.com}           
           \and
           H. Cherifi \at
           Laboratoire Electronique, Informatique et Image (Le2i) UMR 6306 CNRS\\
           University of Burgundy, Dijon, France.\\
          \email{hocine.cherifi@u-bourgogne.fr}           
}

\date{Received: date / Accepted: date}

\maketitle

\begin{abstract}
Although color is a fundamental feature of human visual perception, it has been largely unexplored in the reduced-reference (RR) image quality assessment (IQA) schemes. In this paper, we propose a natural scene statistic (NSS) method, which efficiently uses this information. It is based on the statistical deviation between the steerable pyramid coefficients of the reference color image and the degraded one. We propose and analyze the multivariate generalized Gaussian distribution (MGGD) to model the underlying statistics. In order to quantify the degradation, we develop and evaluate two measures based respectively on the Geodesic distance between two MGGDs and on the closed-form of the Kullback Leibler divergence. We performed an extensive evaluation of both metrics in various color spaces (RGB, HSV, CIELAB and YCrCb) using the TID 2008 benchmark and the FRTV Phase I validation process. Experimental results demonstrate the effectiveness of the proposed framework to achieve a good consistency with human visual perception. Furthermore, the best configuration is obtained with CIELAB color space associated to KLD deviation measure.
\end{abstract}
\section{Introduction}
\label{intro}
As subjective quality evaluation is time consuming, and cannot be used for real time applications, mathematical models that approximate results of subjective quality assessment are needed. The so called "objective image quality assessment models" can be classified according to the information they use about the original image. Full-reference (FR) methods measure the perceived deviation between the degraded image and the original that is assumed to have perfect quality. While FR methods are based on the knowledge of the reference image, no-reference (NR) methods are designed in order to grade the image quality independently of the reference. As they are usually designed for one or a set of specific distortion features, they are useful  when the types of distortions  are fixed and known, and unlikely to be generalized. Reduced-Reference (RR) methods are designed to predict the perceptual quality of distorted images with only partial information about the reference images. Depending on the amount of information of the original image used, both limiting cases (FR and NR) can be achieved. Therefore, the quantity of information extracted from the original image is a critical parameter of the reduced-reference schemes. Methods based on natural image statistics modeling present a good trade-off between the quality prediction accuracy and the quantity of side information needed. The method proposed originally by Wang \emph{et al.}~\cite{RefC} uses an univariate generalized Gaussian distribution GGD to model steerable pyramids coefficients and the Kullback Leibler divergence (KLD) to quantify the distortion. Motivated by perceptual and statistical issues, Lee \emph{et al.}~\cite{RefD} proposed to substitute to the wavelet representation a divisive normalization transform (DNT). It is built upon a wavelet image decomposition, followed by a divisive normalization stage. They assume a Gaussian scale mixture model (GSM) for the wavelet coefficients distribution. Despite its interesting results, the proposed method is much more complex because of the normalization process. A recent work has been proposed by Soundararajan \emph{et al.}~\cite{RefZ}. The so called "reduced reference entropic difference method" (RRED) measure differences between the entropies of wavelet coefficients of reference and distorted images. This method has demonstrated good performances as compared to the human visual perception. However, a great quantity of information from the original image is needed in order to reach a high quality score. While color information is ubiquitous in our world, these metrics have been designed for gray level images, and there has been little work dealing with reduced reference measures for color images. Decherchi \emph{et al.}~\cite{RefG} introduce a machine-learning framework to predict the quality score. In this two-layer architecture, the first layer performs distortion identification, so that images can be forwarded to the identified sub-modules in the second layer specially trained to predict the visual quality for this specific distortion. Each sub module is an improved version of the basic Extreme Learning Machine (ELM) called the Circular-ELM (C-ELM). Both luminance and chrominance information are used for extracting numerical descriptors to feed the predictor. Comparison has shown that their solution compares satisfactorily with other quality assessment methods proposed in the literature for the four types of distortions of the LIVE benchmark (JPEG compression, JPEG 2000 Compression, White noise and Gaussian Blur). We may note that this approach is far more complex than competing methods~\cite{RefAAA}. Furthermore, the complexity increases drastically with the number of distortion types. Another work from Redi \emph{et al.}~\cite{RefF} use descriptors based on the color correlogram for HSV and YCrCb color representation. The hue channel is used in the HSV and luminance channel in YCrCb color space. They also used a computational intelligence paradigm with two layers. The first uses neural classification to identify the kind of distortion while the second is dedicated to the quality prediction using neural regression machine. The alterations in the color distribution of an image is perceived as the consequence of the occurrence from the distortions.  This approach suffers from the same level of complexity as compared to Decherchi's one. There is another machine learning based method developed by He \emph{et al.}~\cite{RefH}. Features extracted using a color fractal structure model are mapped to visual  quality using the support vector regression. Their method demonstrates good performances on the LIVE Benchmark. However, experiments are limited to a reduced number of distortion types.\\
To our knowledge, when dealing with color images existing reduced reference methods process each color component independently without considering the shared information between these components. In order to overcome this drawback, this paper presents a multi-dimensional model of natural color images distribution. The multivariate generalized Gaussian distribution (MGGD) is used as the statistical model of steerable-pyramid coefficients (a redundant transform of wavelets family)~\cite{simoncelli92}. This choice is justified by the non-Gaussianity of the coefficients. This model is used in a RR scheme for color image quality assessment presented in a previous work~\cite{RefW}. As the MGGD model has demonstrated promising results when considering the RGB color representation, we extend our investigation into two directions. First of all, we explore the impact of various color representations on the image quality assessment process. We also investigate the influence of the distortion distance between the statistics of the original and the processed image.\\
 Comparative evaluation is performed with the extension to color of reduced-reference methods based on natural image statistics modeling: Wavelet-domain Natural Image Statistic Model (WNISM) \cite{RefAS} and Reduced Reference Entropic Differencing Index (RRED) \cite{RefZ}.\\
This paper is organized as follows. Section \ref{sec:1} presents the proposed RR image quality measures scheme. Section \ref{sec:4} introduces the different elements that are needed in order to implement the overall process. Section \ref{sec:9} is devoted to the experimental setup. The TID 2008 benchmark is presented and the validation protocol is recalled. Section \ref{sec:12} reports the experimental results and finally concluding remarks are presented in section \ref{sec:16}.
\section{Proposed scheme}
\label{sec:1}
\begin{figure}
\centering
\includegraphics[width=4.5in]{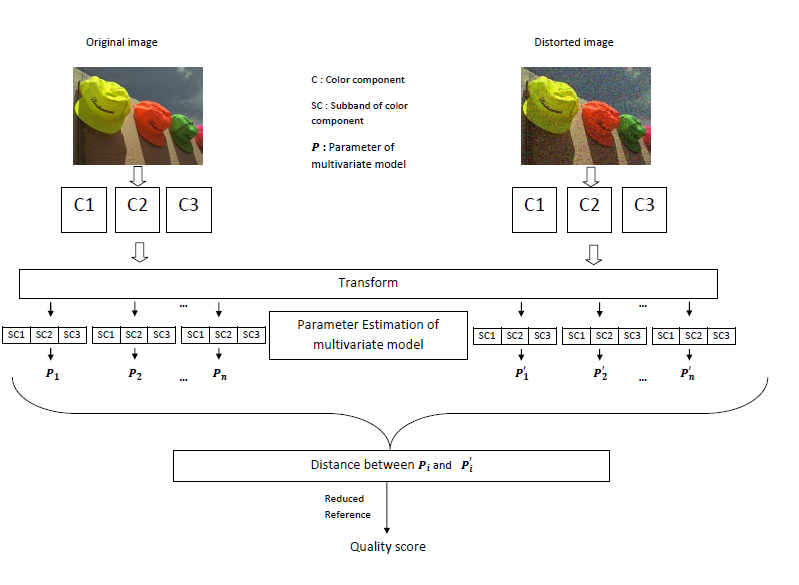}
\caption{Deployment of the reduced reference method in color space with a multivariate model.}
\label{fig:5}
\end{figure}

In this section, we present the framework of the  proposed method based on statistical modeling of color components.
If we consider the general framework of natural image statistics RRIQA (Reduced Reference Image Quality Assessment) scheme, the feature extraction process transforms the image in a suitable representation from which statistics are extracted. At the receiver side, the same feature is extracted and compared to assess the visual quality. The deployment scheme of the proposed RRIQA measure is given in figure \ref{fig:5}. First, eventually, a color transformation of both original and distorted images is used to get the three color components. Then, a spatial frequency domain representation is computed for each color component.  Based on a multivariate distribution model estimates of the distribution, parameters ($P$) are computed on the three color sub-bands of the same decomposition level. Finally, a dissimilarity distance is used in order to compute the distortion measure between the two images. In this context, quality scores are strongly dependent on several parameters: the color space representation, the spatio-frequency transformation, the underlying distribution model together with the estimates of its parameters, and the adequacy of the dissimilarity measure. Furthermore, one cannot neglect the influence of the benchmark and the validation protocol used to predict the subjective quality from objective measurements. In the next section, we introduce and justify the retained solutions for each element of this general scheme.
\section{Methods}
\label{sec:4}
This section gives an overview of the various elements of the quality assessment scheme. We start with a brief description of the four color spaces under investigation (RGB, HSV, CIELAB, and YCrCb). Then, we recall the mathematical description and the parameter estimation of the MGGD model. Finally, we define the overall distortion measure, which is expressed in terms of the proposed dissimilarity measurements (Kullback Leibler divergence or Geodesic distance).
\subsection{Color spaces}
\label{sec:5}
In order to assess the influence of color on the quality, we choose to concentrate on four broadly used color spaces: three device dependent (RGB, HSV, and YCrCb), and one device independent (CIELAB). The basic RGB color space is selected due to its simplicity and suitability. It uses the additive color mixing for the three components red, green, and blue~\cite{RefAM}. The HSV color space is retained because it is very intuitive in order to specify colors~\cite{RefG,RefF,RefH}. It breaks the color space into hue (or color), saturation, and value (or lightness). For this reason it is used in many image-processing applications. YCrCb separates RGB into luminance and chrominance information. Y is the luma, which is very similar to luminance. Cr is the component of red to green, and Cb is the component of blue to yellow. It is a standard useful in analog television and digital video transmission. Indeed, it has been designed to be efficient at encoding RGB values so they consume less space while retaining the full perceptual value. While these color spaces are perceptually non–linear, CIELAB has been designed to accurately map color perception. In other words, a unit change stimulus in CIELAB produce the same change in perception wherever it is applied. The three parameters that define the space are Luminance, red to green, and blue to yellow. In general, to compute CIELAB components from RGB, it is necessary to apply a complete calibration or at least an estimation method~\cite{RefAN}. In the following, we consider here only a standard transformation from the CIEXYZ color space to get an approximation of CIELAB color space.
\subsection{Steerable Pyramid transform}
\label{sec:6}
In order to simulate the functionality of the early human visual system (HVS) components, perceptual image quality assessment methods usually involve a channel decomposition process that transforms the image into different spatial frequency as well as orientation selective sub-bands. This process can be implemented using various multi-scale methods such as wavelet transform, Laplacian pyramid transform and contourlet transform. In this work, we restrict our attention to the steerable pyramid transform because it has proven to be effective in the image quality literature~\cite{RefC,RefD,RefZ}. Furthermore, evaluating the impact of the transformation used on the quality method is beyond the scope of this paper~\cite{RefW}. Nevertheless, it should be noted there is still room for improvement at this level.
\subsection{Statistical Models}
\label{sec:7}
In this work, we consider a particular case of the multivariate generalized Gaussian distribution
introduced by Kotz~\cite{RefO}. It is defined by the following expression:
\begin{equation}
\begin{split}
f(x|\mu,\Sigma,\beta)=\frac{\Gamma(\frac{m}{2})}{\pi^{\frac{m}{2}}\Gamma(\frac{m}{2\beta})2^{\frac{m}{2\beta}}}\frac{\beta}{\left|\Sigma\right|^{\frac{1}{2}}}\\
\times\exp\left\{-\frac{1}{2}\left[(x-\mu)'\Sigma^{-1}(x-\mu)\right]^{\beta}\right\},
 \label{eq:1}
\end{split}
\end{equation}
 Where $m$ is the dimensionality of the probability space ($m = 3$ for color space), $\beta$ is the shape parameter to control the peakedness of the distribution and the heaviness of its tails, $\mu$ is the expectation vector and $\Sigma$ is the covariance matrix of color components for each sub-band. Note that the multivariate generalized Gaussian is also sometimes called the multivariate exponential power distribution.\\
In the case of $\beta=1$, the MGGD becomes a multivariate Gaussian distribution MGD with the following expression:
 \begin{equation}
 f(x|\mu,\Sigma)=\frac{1}{\left(2\pi\right)^{\frac{m}{2}}{\left|\Sigma\right|^{\frac{1}{2}}}}\exp\left\{-\frac{1}{2}\left[(x-\mu)'\Sigma^{-1}(x-\mu)\right]\right\},
 \end{equation}
To estimate the MGGD parameters, we use the maximum likelihood method. It involves setting the differential to zero of the logarithm of $f$ in equation \ref{eq:1}. Arranging the wavelet coefficients for a single sub-band in $n$ three-dimensional column vectors $x_{i}(i=1,.......,n)$, we get the following equations for $\Sigma$ and $\beta$, respectively.
\begin{equation}
\Sigma=\frac{\beta}{n}\sum_{i=1}^{n}u^{\beta-1}x_{i}x{}_{i}^{'}
\end{equation}
\begin{equation}
\sum_{i=1}^{n}\left\{\frac{\beta}{2}\ln\left(u\right)u^{\beta}-\frac{p}{2\beta}\left[\ln\left(2\right)+\Psi\left(\frac{p}{2\beta}\right)\right]-1\right\}=0
\end{equation}
Here $\psi$ denotes the digamma function and $u$ is defined by $x'\Sigma^{-1}x$. These equations are solved recursively.\\
\begin{table}[!h]
\caption{Results with Kolmogorov-Smirnov goodness-of-fit test of a trivariate MGGD model for the two images in TID 2008.}
\label{tab:2}
 \centering %
\begin{tabular}{lllll}
\hline\noalign{\smallskip}
& \multicolumn{2}{c}{MGGD} & \multicolumn{2}{c}{MGD}\\
&  {(I01)} & {(I03)}&  {(I01)} & {(I03)}\\
\noalign{\smallskip}\hline\noalign{\smallskip}
{RGB}&{{$0.069$}}&{{$0.059$}}&{{$0.094$}}&{{$0.078$}}\\
{HSV} & {{$0.082 $}} &{{$0.076$}}& {{$0.155 $}} &{{$0.122$}}\\
{CIELAB}&{{$0.098$}}&{{$0.088$}}&{{$0.247$}}&{{$0.189$}}\\
{YCrCb}&{{$0.093$}}&{{$0.082$}}&{{$0.231$}}&{{$0.180$}}\\
\noalign{\smallskip}\hline
\end{tabular}
\end{table}
To check the adequacy of the model to color image data, we conducted a set of experiments comparing the multivariate generalized Gaussian distribution with the multivariate Gaussian distribution. \\
Table \ref{tab:2} reports typical values of the  Kolmogorov-Smirnov (KS) goodness-of-fit test for the two alternative hypotheses. Note that the smaller the KS value, the better the model fit with the empirical distribution. According to these results, the fits are quite satisfactory and support the use of MGGD for modeling steerable pyramids coefficients. We can notice that the adequacy of the model depends of the color space representation. It is very effective for the RGB color space which is the most correlated one. Differences are less pronounced between HSV, YCrCb and Lab, even if results are better for the former.
\subsection{Distortion measure}
\label{sec:8}
At the receiver side, we use the features sent from the reference image to compute the distortion measure. Here, we propose to exploit conjointly the dependencies between the sub-bands and the dependencies between the color components. To get the overall difference between images of $L$ sub-bands, we use the following equation:
\begin{equation}
D=\text{\ensuremath{\sum}}_{i=1}^{L}d_{i}\left(p_{1}\left(x|\theta_{1}\right)||p_{2}\left(x|\theta_{2}\right)\right)
 \end{equation}
Where $p_1(x|\theta_{1})$ represents the distribution of the original image, while $p_2(x|\theta_{2})$ represents the distribution of the degraded image. The parameters vector $\theta_1$ is  sent with the reference image in order to estimate $p_1(x|\theta_{1})$, and $\theta_2$ is computed at the receiver side to estimate $p_2(x|\theta_{2})$.
The overall quality measure is defined as follows:
\begin{equation}\label{eq:5}
Q=\log{_{2}}\left(1+\frac{1}{D_{0}}D\right)
\end{equation}
$D_{0}$ is a constant to control the magnitude of the distortion measure. In the experiments, we set its value to 0.1. The logarithmic function is used to reduce the difference between high values and low values of $D$, so that we can have values in the same order of magnitude.\\
In order to quantify the difference $d_{i}$ between $ p_{1}\left(x|\theta_{1}\right)$ and $ p_{2}\left(x|\theta_{2}\right)$, a dissimilarity measure is required. In the literature, several distances exist. We choose to investigate two measures, the Kullback-Leibler divergence KLD and the geodesic distance GD.
\paragraph{\textbf{Kullback-Leibler divergence between two MGGDs:}}
 An approximation of the analytic expression of the KLD between two multivariate zero-mean GGDs parameterized by $\left(\beta_{1},\Sigma_{1}\right)$ and $\left(\beta_{2},\Sigma_{2}\right)$ has been proposed by Verdoolaege \emph{et al.}~\cite{RefR}. This measure denoted $D_{KL}$ is defined by :
\begin{equation}
\begin{split}
D_{KL}\left(\beta_{1},\Sigma_{1}||\beta_{2},\Sigma_{2}\right)=\ln\left[ \frac{\Gamma\left(\frac{1}{\beta_{2}}\right)}{\Gamma\left(\frac{1}{\beta_{1}}\right)}
2^{\left(\frac{1}{\beta_{2}}-\frac{1}{\beta_{1}}\right)}\left(\frac{|\Sigma_{2}|}
{|\Sigma_{1}|}\right)^{\frac{1}{2}}\frac{\beta_{1}}{\beta_{2}}\right] \\
-\left(\frac{1}{\beta_{1}}\right)+\Biggl[2^{\left(\frac{\beta_{2}}{\beta_{1}}-1\right)}
\frac{\Gamma\left(\frac{\beta_{2}+1}{\beta_{1}}\right)}{\Gamma\left(\frac{1}{\beta_{1}}\right)}\\
\times\left(\frac{\gamma_{1}+\gamma_{2}}{2} \right)^{\beta_{2}} F_{1}\left(\frac{1-\beta_{2}}{2},\frac{-\beta_{2}}{2};1;A^{2} \right)\Biggr].
\end{split}
\end{equation}
Where, $F_{1}(.,.;.;.)$ represents the Gauss hyper-geometric function (Abramowitz and Stegun, 1965)~\cite{RefS} which may be tabulated for $−1 < A < 1$ and for realistic values of $\beta$. $\gamma_{i}=\left(\lambda^{i}_2\right)^{-1}$, with $i=1,2,$ and $\lambda^{i}_2$ is the eigenvalues of $\Sigma_{2}^{-1}\Sigma_{1}$ while $A \equiv\frac{\gamma_{1}-\gamma_{2}}{\gamma_{1}+\gamma_{2}} $.
\paragraph{\textbf{Geodesic distance:}}
In particular, the geodesic distance between two MGGDs with fixed shape parameter $\beta$ and with covariance matrices $\Sigma_{1}$ and $\Sigma_{2}$ as defined by Verdoolaedge \emph{et al.}~\cite{RefN} is given by:
\begin{equation}
\begin{split}
D_{geo}\left(\Sigma_{1}||\Sigma_{2}\right)=\frac{1}{|H|^{2p}}\Biggl[\left(3b_{GG}-\frac{1}{4}\right)
\text{\ensuremath{\sum}}_{i=1}^{p}\left(\ln\lambda_{i}\right)^{2} \\
+2\left(b_{GG}-\frac{1}{4}\right)\text{\ensuremath{\sum}}_{i<j}^{p}\left(\ln\lambda_{i}\right)\left(\ln\lambda_{j}\right)\Biggr]^{\frac{1}{2}}.
\end{split}
\end{equation}
With, $H$ is a matrix that diagonalizes $\Sigma_{1}$ and $\Sigma_{2}$ simultaneously so that $\Sigma_{1}$ is transformed into the unit matrix, and $\lambda_{i} (i=1,...,p)$ are the eigenvalues of $\Sigma_{1}^{-1}\Sigma_{2}$. In addition, $b_{GG}$ is defined by $b_{GG}\equiv \frac{1}{4}\frac{p+2\beta}{p+2}$. We denote that $p$ is the number of the eigenvalues, and $\beta$ is the shape parameter.
\section{Experimental setup}
\label{sec:9}
\subsection{Datasets}
\label{sec:10}
To test the performances of quality algorithms, a dataset of distorted images graded by human observers is needed. As the evaluation process is greatly influenced by the semantic value of the images, care must be taken when choosing the dataset. It must contain images that reflect adequate diversity in image content and generated distortions should reflect a broad range of image impairments. To comply with this argument, the proposed method has been tested and validated using two datasets (TID 2008 dataset, LIVE dataset) specially designed for quality metrics evaluation. The LIVE dataset contains 29 high-resolution reference image altered with five distortion types (JPEG2000, JPEG, White noise, Gaussian blur, and Fast fading). Despite the fact that it does not cover a broad range of distortions, we retained it because it uses to be the only benchmark, and it has been widely used in the literature. Nevertheless, we do not report the results with this benchmark for lack of space. Furthermore, there is no major differences with the most recent benchmark test publicly available TID 2008~\cite{RefT}. It contains 25 reference images, and 1700 distorted images (25 reference images with 17 types of distortions and 4 levels of distortions). These distortions type are: Additive Gaussian noise, Additive noise in color components, Spatially correlated noise, Masked noise, High frequency noise, Impulse noise, Quantization noise, Gaussian blur, Image denoising, JPEG compression, JPEG2000 compression, JPEG transmission errors, JPEG2000 transmission errors, Non eccentricity pattern noise, Local block-wise distortions of different intensity, Mean shift (intensity shift) and Contrast change. The quality of each image in TID 2008 has been graded by the Mean Opinion Score (MOS) derived from psychophysical experiments.
\subsection{Validation protocol}
\label{sec:11}
 We use a non-linear function proposed by the Video Quality Expert Group (VQEG) Phase I FR-TV with five parameters\cite{RefV}. The expression of the quality score which is the predicted MOS is given by:
\begin{equation}
\label{eq:12}
DMOS_{p}=\beta_{1}\mathrm{logistic}\left(\beta_{2},Q-\beta_{3}\right)+\beta_{4}Q+\beta_{5}.
\end{equation}
Where the vector $(\beta_{1},\beta_{2},\beta_{3},\beta_{4},\beta_{5})$ is estimated thanks to \emph{fminsearch} function in the optimization Toolbox of Matlab, and the logistic function is expressed by :
\begin{equation}
\mathrm{logistic}\left(\tau,Q\right)=\frac{1}{2}-\frac{1}{1+\exp\left(\tau Q\right)}.
\end{equation}
Where Q is the overall measure in equation \ref{eq:5}.\\
To evaluate the relevance of a quality metric, we use two measures. The prediction accuracy is measured by the Pearson's linear correlation coefficient PLCC, while the  prediction monotonicity  is quantified by the Spearman rank-order coefficient SRCC. These measures are defined as follows:
\begin{equation}
PLCC=\frac {\sum_{i=1}^{N}(DMOS(i)-\bar{DMOS})(DMOS_p(i)-\bar{DMOS_p})}{\sqrt {\sum_{i=1}^{N}(DMOS(i)-\bar{DMOS})^{2}} \sqrt{\sum_{i=1}^{N}(DMOS_p(i)-\bar{DMOS_p})^{2}}},
\end{equation}
\begin{equation}
SRCC=1-\frac{6\sum_{i=1}^{^{N}}(rank(DMOS(i))-rank(DMOS_p(i)))^{2}}{N(N^{2}-1)},
\end{equation}
Where $i$ denotes the index of the image sample and $N$ is the number of samples. $DMOS$ presents here the subjective scores, and $DMOS_{p}$ presents the predicted MOS defined in eq. \ref{eq:12}.
\section{Experimental results}
\label{sec:12}
This section reports the results of a detailed comparison of the various combinations of color spaces, and dissimilarity measures used in the proposed scheme. Experiments have been performed using the TID 2008 benchmark because it contains a broad spectrum of distortions. Performances are evaluated in terms of accuracy and monotonicity. At first, we concentrate on the influence of the color space on performances. Then, we examine the impact of the dissimilarity measures. Finally, we compare the most efficient combination of parameters with competing methods. In order to make a fair comparison, the number of extracted features from the original image by each method and the computational time are also reported.
\subsection{Color spaces evaluation}
\label{sec:13}
Let's start our comparison with the accuracy of the different color spaces. Table \ref{tab:3} reports the PLCC values for each distortion type and the two distance measures (KLD, GD) using the TID 2008 dataset. Furthermore, the last line with the distortion type labeled "all" gives the PLCC value computed with the entire dataset. In each case, The PLCC value of the most efficient color space is reported in bold.\\
When the geodesic distance is used, the worst scores are obtained for the distortion induced by the contrast change and this is true for all the color spaces under study. In this case, we can conclude that the measure fails completely. Substituting the KLD to GD allows a significant improvement for HSV, CIELAB, and YCrCb but it is not the case for RGB, whose accuracy deteriorates. RGB is the most efficient for four artifacts (Gaussian blur, JPEG compression, Mean shift, Contrast change).  HSV is the most efficient in three cases (Additive noise in color components, Spatially correlated noise, impulse noise). It also outperforms the other color spaces when the entire data set is used. YCrCb gives the best accuracy value for five distortion types (Masked noise, High frequency noise, Image denoising, Non eccentricity pattern noise, Local block-wise distortions of different intensity). For the five remaining distortion types, CIELAB dominates. \\
When the KLD is used, the worst scores are obtained for the Non eccentricity pattern noise with PLCC values below 0.6. In this case, again results are very consistent for all the color spaces. In other words, the measure is inefficient for this type of distortion unless one uses GD instead of KLD. Note also that except for the HSV color space, the intensity shift distortion is not handled accurately by the measure. KLD associated to RGB seems to be the worst combination. Indeed, in this situation RGB is never one of the most efficient color spaces. If we consider the number of distortion types for which a color space ranks first, CIELAB outperforms its competitors. Indeed, it is the most performing in eight situations. It is followed by HSV with seven best scores. YCrCb and RGB are far behind with respectively three and zero best scores. Nevertheless, it is worth noting that HSV remains the most performing in the case where the entire dataset is considered. According to these results, it appears that we can give a slight advantage to CIELAB in term of accuracy for a great number of distortion types, and an advantage to HSV in term of overall robustness. Furthermore, if RGB must be used, it is better to associate it with GD rather than the KLD.\\
\begin{table}[!h]
\caption{The Pearson's linear correlation coefficients (PLCC) in TID 2008.}
\label{tab:3}
 \centering %
\begin{tabular}{lllllllll}
\hline\noalign{\smallskip}
 & \multicolumn{4}{c}{Geodesic Distance (GD)} & \multicolumn{4}{c}{Kullback Leibler (KLD)}\\
{Distortion type } & {RGB} & {HSV} & {CIELAB} & {YCrCb} & {RGB} & {HSV} & {CIELAB} & {YCrCb}\\
\noalign{\smallskip}\hline\noalign{\smallskip}
{Additive Gaussian noise} &{0.785} &{0.783} &{\textbf{0.878}} & {0.793} & {0.829} & {\textbf{0.852}} & {0.831}&{0.793}\\
{Additive noise in} &{0.801} &{\textbf{0.839}} & {0.802} & {0.800} & {0.836} & {0.845} & {0.855}&{\textbf{0.869}}\\
{Color components ...} & & & &  & & &&\\
{Spatially correlated noise} &{0.841} &{\textbf{0.914}} &{0.826} & {0.805} & {0.873} & {\textbf{0.917}} & {0.892}&{0.821}\\
{Masked noise} &{0.840} &{0.823} & {0.834}& {\textbf{0.872}} & {0.842} & {0.884} & {\textbf{0.897}}&{0.870}\\
{High frequency noise} &{0.857} &{0.870} & {0.846} & {\textbf{0.893}} & {0.871} & {0.885} & {0.877}&{\textbf{0.896}}\\
{Impulse noise} &{0.813} &{\textbf{0.839}} & {0.746} & {0.747} & {0.829} & {\textbf{0.848}} & {0.769}&{0.771}\\
{Quantization noise} & {0.803}&{0.801} & {\textbf{0.828}}& {0.812} & {0.824} & {0.833} & {\textbf{0.855}}&{0.835}\\
{Gaussian blur} & {\textbf{0.948}}&{0.937} &{0.942} & {0.873} & {0.923} & {0.817} & {\textbf{0.924}}&{0.820}\\
{Image denoising} &{0.824} &{0.834} & {0.883} & {\textbf{0.887}} & {0.903} & {0.916} & {\textbf{0.941}}&{0.940}\\
{JPEG compression} &{\textbf{0.865}} &{0.822} & {0.810} & {0.807} & {0.830} & {0.870} & {\textbf{0.873}}&{0.871}\\
{JPEG2000 compression} &{0.879} &{0.889} & {\textbf{0.894}} & {0.886} & {0.904} & {0.919} & {\textbf{0.944}}&{0.942}\\
{JPEG transmission compression} &{0.827} &{0.868} & {\textbf{0.894}} & {0.810} & {0.852} & {0.837} & {\textbf{0.895}}&{0.892}\\
{JPEG2000 transmission compression} &{0.737} &{0.779} & {\textbf{0.846}} & {0.780} & {0.712} & {0.796} & {\textbf{0.856}}&{0.821}\\
{Non eccentricity pattern noise} &{0.755} &{0.759} & {0.790} & {\textbf{0.793}} & {0.544} & {\textbf{0.596}} & {0.555}&{0.542}\\
{Local block-wise ...} &{0.796} &{0.832}& {0.806} & {\textbf{0.842}} & {0.807} & {0.823} & {0.841}&{\textbf{0.861}}\\
{Mean shift (intensity shift)} &{\textbf{0.874}} &{0.869} & {0.714} & {0.732} & {0.576} & {\textbf{0.886}} & {0.639}&{0.686}\\
{Contrast change} &{\textbf{0.653}} &{0.635} & {0.556} & {0.593} & {0.528} & {\textbf{0.827}} & {0.745}&{0.752}\\
{All} &{0.817} & {\textbf{0.829}} & {0.817}& {0.807} & {0.793} & {\textbf{0.844}} & {0.835}&{0.822}\\
\noalign{\smallskip}\hline
\end{tabular}
\end{table}
Let's now turn to the monotonicity as a comparison criterion of the different color spaces. Table~\ref{tab:5} reports the results obtained with the TID 2008 dataset. For the Geodesic distance, according to the SRCC values, RGB is the most efficient color space for five types of distortion (masked noise, high frequency noise, JPEG compression, JPEG2000 transmission errors, Local block-wise). HSV and YCrCb outperform the other color spaces in two cases: additive Gaussian noise and quantization noise for HSV, and spatially correlated noise and Impulse noise for YCrCb. For the eight remaining distortion types, CIELAB surpasses them (additive noise in color components, Gaussian blur, image denoising, JPEG2000 compression, JPEG transmission errors, Non eccentricity pattern noise, Mean shift, Contrast change). This is also the case when the entire dataset is used.\\
For the KLD, CIELAB remains the most efficient for seven distortions types (quantization noise, Gaussian blur, image denoising, JPEG2000 compression, JPEG transmission errors, JPEG2000 transmission errors, Contrast change) and when the entire dataset is used. The HSV performance increases. Indeed, it takes the first rank for six distortion types (additive Gaussian noise, additive noise in color components, Impulse noise, JPEG compression, non eccentricity pattern noise, Mean shift). RGB performance deteriorates. It remains with the highest scores for only two distortion types (masked noise, high frequency noise). Again, YCrCb exceeds the three other color spaces only in two distortion types (Spatially correlated noise, Local block-wise). We note also that both KLD and GD fail for the mean shift and contrast change artifact with a monotonicity coefficient under or close to 0.7.\\
To summarize, the CIELAB color space is the best compromise, followed by HSV. Note that, even if it is almost always behind CIELAB, YCrCb is nevertheless very close to it. Furthermore, remember that CIELAB is obtained by a standard transformation as stated in section 3.1. This issue can have an important effect in the related result in the comparison of color spaces. Nevertheless, we believe that it can be considered as a worst case situation for CIELAB.\\
\begin{table}[!h]
\caption{The Spearman rank correlation coefficients (SRCC) for TID 2008 database.}
\label{tab:5}
 \centering %
\begin{tabular}{lllllllll}
\hline\noalign{\smallskip}
 & \multicolumn{4}{c}{Geodesic Distance (GD)} & \multicolumn{4}{c}{Kullback Leibler (KLD)}\\
{Distortion type } & {RGB} & {HSV} & {CIELAB} & {YCrCb} & {RGB} & {HSV} & {CIELAB}&{YCrCb}\\
\noalign{\smallskip}\hline\noalign{\smallskip}
{Additive Gaussian noise} & {0.792} & {\textbf{0.837}} & {0.785} & {0.746} & {0.837} & {\textbf{0.865}} & {0.852}& {0.843} \\
{Additive noise in} & {0.834} & {0.887} & {\textbf{0.896}} & {0.842} & {0.872} & {\textbf{0.891}} & {0.886}& {0.871} \\
{Color components ...} & & & & &  & & & \\
{Spatially correlated noise} & {0.852} & {0.801} & {0.832} & {\textbf{0.867}} & {0.885} & {0.894} & {0.890}& {\textbf{0.896}} \\
{Masked noise} & {\textbf{0.905}} & {0.830} & {0.834} & {0.850} & {\textbf{0.942}} & {0.856} &{0.851} & {0.864} \\
{High frequency noise} &{\textbf{0.888}} & {0.885} & {0.837} & {0.843} & {\textbf{0.909}} & {0.834} & {0.872}& {0.877} \\
{Impulse noise} &{0.815} &{0.836} & {0.844} & {\textbf{0.845}}  & {0.839} & {\textbf{0.861}} & {0.847}& {0.845} \\
{Quantization noise} &{0.828} &{\textbf{0.881}} & {0.869} & {0.863} & {0.859} & {0.882} & {\textbf{0.911}}& {0.902} \\
{Gaussian blur} & {0.938} & {0.817} & {\textbf{0.951}} & {0.949} & {0.930} & {0.826} &{\textbf{0.952}} & {0.947} \\
{Image denoising} &{0.840} &{0.938} & {\textbf{0.965}} & {0.961} & {0.942} & {0.949} & {\textbf{0.973}}& {0.966} \\
{JPEG compression} &{\textbf{0.893}} &{0.831} &{0.851} & {0.853} & {0.842} & {\textbf{0.865}} & {0.853}& {0.854} \\
{JPEG2000 compression} & {0.825}&{0.801} & {\textbf{0.880}} & {0.871} & {0.932} & {0.898} & {\textbf{0.942}}& {0.937} \\
{JPEG transmission errors} & {0.842} & {0.877} & {\textbf{0.885}} & {0.880} & {0.878} & {0.899} &{\textbf{0.921}}& {0.915} \\
{JPEG2000 transmission errors} & {\textbf{0.828}} & {0.827} & {0.802} & {0.803} & {0.867} & {0.836} &{\textbf{0.884}}& {0.879} \\
{Non eccentricity pattern noise} &{0.784} & {0.791} & {\textbf{0.800}} & {0.795} & {0.746} & {\textbf{0.832}} & {0.714}& {0.725}\\
{Local block-wise ...} & {\textbf{0.876}} & {0.866} & {0.822} & {0.824} & {0.819} & {0.850} & {0.879}& {\textbf{0.882}}\\
{Mean shift (intensity shift)} & {0.691} & {0.698} & {\textbf{0.714}} & {0.703} & {0.619} & {\textbf{0.755}} & {0.645}& {0.651}\\
{Contrast change} & {0.667} & {0.688} & {\textbf{0.697}} & {0.691} & {0.662} & {0.685} & {\textbf{0.767}}& {0.637}\\
{All} &{0.829} & {0.829} & {\textbf{0.839}} & {0.834} & {0.846} & {0.852} & {\textbf{0.861}}& {0.852}\\
\noalign{\smallskip}\hline
\end{tabular}
\end{table}
\subsection{Comparison between the Geodesic Distance and the Kullback Leibler Divergence}
\label{sec:14}
The Kullback-Leibler divergence is the most popular dissimilarity measure used in image quality assessment schemes. We also propose to test the Geodesic distance due to its suitability to the MGGD model~\cite{RefN}. Let's return to accuracy and monotonicity for TID 2008 dataset as presented in Tables \ref{tab:3} and \ref{tab:5}. If we compare the accuracy values independently of the color space, the KLD is the most effective as compared to geodesic distance except in three situations (Additive Gaussian noise, Gaussian blur and Non eccentricity pattern noise). When we compare the monotonicity values the results are strongly correlated with prediction accuracy. In this case, the KLD dominates the situation, with a higher performance except for JPEG compression distortion and additive noise in color component.\\
 \begin{table}[!h]
\caption{Comparison with RRED and WNISM methods in CIELAB color space and PLCC for TID 2008 database.}
\label{tab:7}
 \centering %
\begin{tabular}{lllll}
\hline\noalign{\smallskip}
{Distortion type } & {{WNISM~\cite{RefAS}}} &{{$RRED_{16}^{M16}$~\cite{RefZ}}}& {GD}& {KLD}\\
\noalign{\smallskip}\hline\noalign{\smallskip}
{Additive Gaussian noise} & {0.799}& {0.813} & {\textbf{0.878}} & {0.831}\\
{Additive noise in} & {0.813} & {0.834} & {0.802} & {\textbf{0.855}}\\
{color components ...} &  & & \\
{Spatially correlated noise}& {0.825}& {0.845} & {0.826} & {\textbf{0.892}}\\
{Masked noise}& {0.611}& {0.833} & {0.834} & {\textbf{0.897}} \\
{high frequency noise}& {0.687} & {\textbf{0.897}}& {0.846} & {0.877}\\
{Impulse noise}& {0.586}& {0.732} & {0.746} & {\textbf{0.769}}\\
{Quantization noise}& {0.587}& {0.833} & {0.828} & {\textbf{0.855}}\\
{Gaussian blur}& {0.833}& {\textbf{0.957}} & {0.942} & {0.924} \\
{Image denoising}& {0.602}& {\textbf{0.956}} & {0.883} & {0.941}\\
{JPEG compression}& {0.605}&{\textbf{0.939}} & {0.810} & {0.873}\\
{JPEG2000 compression}& {0.865}& {\textbf{0.961}}& {0.894} & {0.944}\\
{JPEG transmission errors}& {0.702}& {0.868} & {0.894} & {\textbf{0.895}}\\
{JPEG2000 transmission errors}& {0.478}& {0.725} & {0.846} & {\textbf{0.856}} \\
{Non eccentricity pattern noise}& {0.621}& {0.743} & {\textbf{0.790}} & {0.555} \\
{Local block-wise ...}& {0.412}& {0.802} & {0.806} & {\textbf{0.841}}\\
{Mean shift (intensity shift)}& {0.714}& {0.523} & {\textbf{0.714}} & {0.639}\\
{Contrast change}& {0.559}& {0.523} & {0.556} & {\textbf{0.745}}\\
{All} &{0.665}&{0.743} & {0.817} & {\textbf{0.835}}\\

\noalign{\smallskip}\hline
\end{tabular}

\end{table}
\begin{table}[!h]
\caption{Comparison with RRED and WNISM methods in CIELAB color space and SRCC for TID 2008 database.}
\label{tab:8}
\vspace{1.5ex}
 \centering %
\begin{tabular}{lllll}
\hline\noalign{\smallskip}
{Distortion type } &{WNISM~\cite{RefAS}}&{{$RRED_{16}^{M16}$~\cite{RefZ}}}& {GD}& {KLD}\\
\noalign{\smallskip}\hline\noalign{\smallskip}
{Additive Gaussian noise} &{0.720}&{0.832}& {0.785} & {\textbf{0.852}}\\
{Additive noise in} &{0.812}&{0.861}& {\textbf{0.896}} & {0.886} \\
{color components ...} & & & \\
{Spatially correlated noise}&{0.804}&{0.851} & {0.832} & {\textbf{0.890}} \\
{Masked noise} &{0.610}&{0.841} & {0.834} & {\textbf{0.851}} \\
{high frequency noise}&{0.712}&{\textbf{0.905}} &{0.837} & {0.872} \\
{Impulse noise}&{0.564}&{0.753} &{\textbf{0.844}} &{0.839} \\
{Quantization noise}&{0.641}&{0.843} &{0.869} &{\textbf{0.911}} \\
{Gaussian blur} &{0.602}&{\textbf{0.959}} & {0.951} & {0.952} \\
{Image denoising}&{0.863}&{0.951} &{0.965} &{\textbf{0.973}} \\
{JPEG compression}&{0.605}&{\textbf{0.941}} &{0.851} &{0.853} \\
{JPEG2000 compression}&{0.598}&{\textbf{0.972}} & {0.880}&{0.942} \\
{JPEG transmission errors}&{0.598}&{0.874} & {0.885} & {\textbf{0.921}} \\
{JPEG2000 transmission errors}&{0.780}&{0.752} & {0.802} & {\textbf{0.884}} \\
{Non eccentricity pattern noise}&{0.664}&{0.731} &{\textbf{0.800}} & {0.714} \\
{Local block-wise ...}&{0.432}&{0.841} & {0.822} & {\textbf{0.879}} \\
{Mean shift (intensity shift)}&{0.596}&{0.568} & {\textbf{0.714}} & {0.645} \\
{Contrast change} &{0.315}&{0.512} & {0.697} & {\textbf{0.767}} \\
{All} &{0.642}&{0.823} & {0.839} & {\textbf{0.861}}\\
\noalign{\smallskip}\hline
\end{tabular}
\vspace{1.5ex}

\end{table}

\begin{table}[htb]
\caption{Comparison summary of KLD and GD measures with RRED and WNSIM methods in CIELAB color space.}
\label{tab:11}
\vspace{1.5ex}
 \centering %
\begin{tabular}{lllll}
\hline\noalign{\smallskip}
  &{WNISM~\cite{RefAS}} &{$RRED_{16}^{M16}$~\cite{RefZ}}&{GD} &{KLD}\\
\noalign{\smallskip}\hline\noalign{\smallskip}
{SRCC} & {{0.642}}&{{0.823}}&{{0.839}}&{{\textbf{0.861}}} \\{PLCC} & {{0.665}}&{{0.743}}&{{0.817}}&{\textbf{{0.835}}}\\ {Number of features} & \textbf{{{36}}}&{{3L/36}}&{{48}}&{{48}} \\{Computation time/image(s)} & {{18.5}}&{\textbf{{3.24}}}&{{19.3}}&{{12.3}}\\
\noalign{\smallskip}\hline
\end{tabular}


\end{table}
\subsection{Comparison with alternative methods}
\label{sec:15}
Here, we compare the proposed measure with the most prominent RRIQA methods based on natural image statistics modeling: Wavelet-domain Natural Image Statistic Model (WNISM)~\cite{RefAS} and Reduced Reference Entropic Differencing Index (RRED)~\cite{RefZ}. As these methods have proven good performance for gray-scale images, we performed a trivial extension to color images. To do so, we computed the measures independently on each color component. Finally, the quality metric is calculated by averaging the color components measures. In order to have a fair comparison, we launched several experiments with these two methods, using all the color spaces and the same transform (steerable pyramid transform with six sub-bands). It appears that generally CIELAB outperforms its competitors. Therefore, in the following, we concentrate on the experiments with CIELAB. Table \ref{tab:7} and \ref{tab:8} report respectively the accuracy and the monotonicity values for the proposed measure using the GD distance (abbreviated as GD), the same measure with the KLD distance (abbreviated as KLD), WNISM and RRED for the TID 2008 database. The results clearly demonstrate the superiority of the proposed measure with the KLD distance as compared to its alternatives. Indeed, according to PLCC, it ranks first for nine type of distortion and when the entire dataset is used. RRED gives better results in five situations (high frequency noise, Gaussian blur, image denoising, JPEG and JPEG 2000), while the proposed measure using the GD distance performs better in three cases (Additive Gaussian noise, non eccentricity pattern noise and mean shift). Note that WNISM never ranks first. Results are quite similar when we observe the SRCC values. KLD is better for nine distortions types against four for RRED and GD. However, differences remain small given the wide variety of distortions and their high intensities. The WNISM method remains the less effective.\\
The goal of a reduced reference image quality assessment measure is to provide the best compromise between prediction accuracy, feature numbers and computational time. In order to give an overall comparison of these methods, table~\ref{tab:11} summarizes all the information in order to compare the four methods.\\
We note that the amount of information transmitted by RRED is very high ($3L/36$ with $L$ is the size of the image in pixels). In the proposed measure, the distribution of each sub-band coefficients is modeled by a trivariate MGGD and the covariance matrix is of $3\times3$ size. For each sub-band, the scale and the shape parameters need to be transmitted together with six coefficients of the symmetric covariance matrix. So, the number of features involved in the quality evaluation process is equal to $48$ for six sub-bands. The number of features is lower for WNISM but accuracy and monotonicity also.\\
Besides data rate and performances, we report the execution time of the four algorithms on an INTEL CORE I3, CPUs 3.3 GHz speed and 4 GB RAM. Note that the computation time is obtained by averaging the computation time required for ten randomly selected images from TID 2008 database. RRED is the most efficient with an execution time of $3.24$ seconds per image followed by KLD, which takes four times longer. The computation time for the GD method and WNISM are very close. It is 1.5 times higher as compared to KLD and about 6 times longer than the execution time of RRED.
\section{Conclusion}
\label{sec:16}
In this paper, a new reduced reference method for color image quality assessment is introduced based on the natural image statistics paradigm. In order to take into account the dependencies between the color components, we propose to use the multivariate generalized Gaussian distribution model. In this context, the visual quality degradation can be computed using various color space and distances between the distributions estimated from the original and distorted images. In order to select the best compromise, we conducted an extensive comparative evaluation involving the TID 2008 dataset. Four color spaces (RGB, HSV, CIELAB, and YCrCb) and two distances (Kullback-Leibler divergence or the Geodesic distance) have been investigated. According to the overall evaluation of accuracy and monotonicity, it appears that CIELAB associated to the KLD dissimilarity is the best candidate to get a reliable color image quality measure. Furthermore, it is worth noting that HSV and YCrCb follow closely. The proposed measure has been also compared with the natural extension to color of the well known WNISM and RRED methods. The results demonstrate that the proposed method presents a good trade-off between the number of extracted features, the quality prediction accuracy and the computational time. As a future work, we plan to investigate other models such as the Gaussian Scale Mixture in order to reduce the number of features while maintaining the effectiveness of the proposed scheme. Furthermore, we plan to extend this work to video quality assessment.
\begin{acknowledgements}
This work has been supported by the project CNRS-CNRST STIC 02/2014.
\end{acknowledgements}




\end{document}